\documentclass[11pt,a4paper]{article}
\usepackage[hyperref]{acl2021}
\usepackage{times}
\usepackage{latexsym}

\usepackage{color}
\usepackage{xspace}
\usepackage{booktabs}
\usepackage{graphicx}
\usepackage{amsmath}
\usepackage{amsfonts}
\usepackage{amssymb}
\usepackage{array}
\usepackage{multirow}
\usepackage[caption=false]{subfig}
\usepackage[ruled,vlined]{algorithm2e}
\usepackage{url}
\usepackage{comment}
\usepackage{amsthm}
\usepackage[caption=false]{subfig}

\theoremstyle{definition}
\newtheorem{definition}{Definition}[]

\DeclareMathOperator*{\argmax}{argmax}

\newcommand{\paratitle}[1]{\noindent \textbf{#1}}

\newcommand{\ie}{\emph{i.e.,}\xspace}
\newcommand{\eg}{\emph{e.g.,}\xspace}

\newcommand{\wrt}{\emph{w.r.t.}\xspace}

\newcommand{\unused}{unmatched\xspace}

\def\tinycol{\hskip 5pt}

\makeatletter
\newcommand*{\balancecolsandclearpage}{%
  \close@column@grid
  \clearpage
  \twocolumngrid
}
\makeatother

\usepackage{microtype}

\aclfinalcopy 
\def\aclpaperid{3031} 

\title{CoRI: \textbf{Co}llective \textbf{R}elation \textbf{I}ntegration with Data Augmentation\\for Open Information Extraction}

\author{
Zhengbao Jiang$^{1}$\thanks{~~This work was performed while at Amazon.}, \quad Jialong Han$^2$, \quad Bunyamin Sisman$^2$, \quad Xin Luna Dong$^2$ \\
Language Technologies Institute, Carnegie Mellon University$^1$\\
Amazon$^2$ \\
\texttt{zhengbaj@cs.cmu.edu} \\
\texttt{\{jialongh,bunyamis,lunadong\}@amazon.com}
}

\date{}

\begin{document}
\maketitle
\begin{abstract}
Integrating extracted knowledge from the Web to knowledge graphs (KGs) can facilitate tasks like question answering.
We study relation integration that aims to align free-text relations in subject-relation-object extractions to relations in a target KG.
To address the challenge that free-text relations are ambiguous, previous methods exploit neighbor entities and relations for additional context.
However, the predictions are made independently, which can be mutually inconsistent.
We propose a two-stage \textbf{Co}llective \textbf{R}elation \textbf{I}ntegration (CoRI) model, where the first stage independently makes candidate predictions, and the second stage employs a collective model that accesses all candidate predictions to make globally coherent predictions.
We further improve the collective model with augmented data from the portion of the target KG that is otherwise unused.
Experiment results on two datasets show that CoRI can significantly outperform the baselines, improving AUC from .677 to .748 and from .716 to .780, respectively.
\end{abstract}

\section{Introduction}\label{sec:intro}

With its large volume, the Web has been a major resource for knowledge extraction.
Open information extraction (open IE;~\citealt{sekine-2006-openie,banko-2007-openie}) is a prominent approach that harvests subject-relation-object \emph{extractions} in free text without assuming a predefined set of relations.
One way to empower downstream applications like question answering is to integrate those free-text extractions into a knowledge graph (KG), \eg Freebase.
\emph{Relation integration} is the first step to integrate those extractions, where their free-text relations (\ie \emph{source relations}) are normalized to relations in the target KG (\ie \emph{target relations}).
Only after relation integration can entity linking proceed to resolve the free-text subjects and objects to their canonical entities in the target KG.

\begin{figure}
\centering
\includegraphics[width=\columnwidth, clip, keepaspectratio]{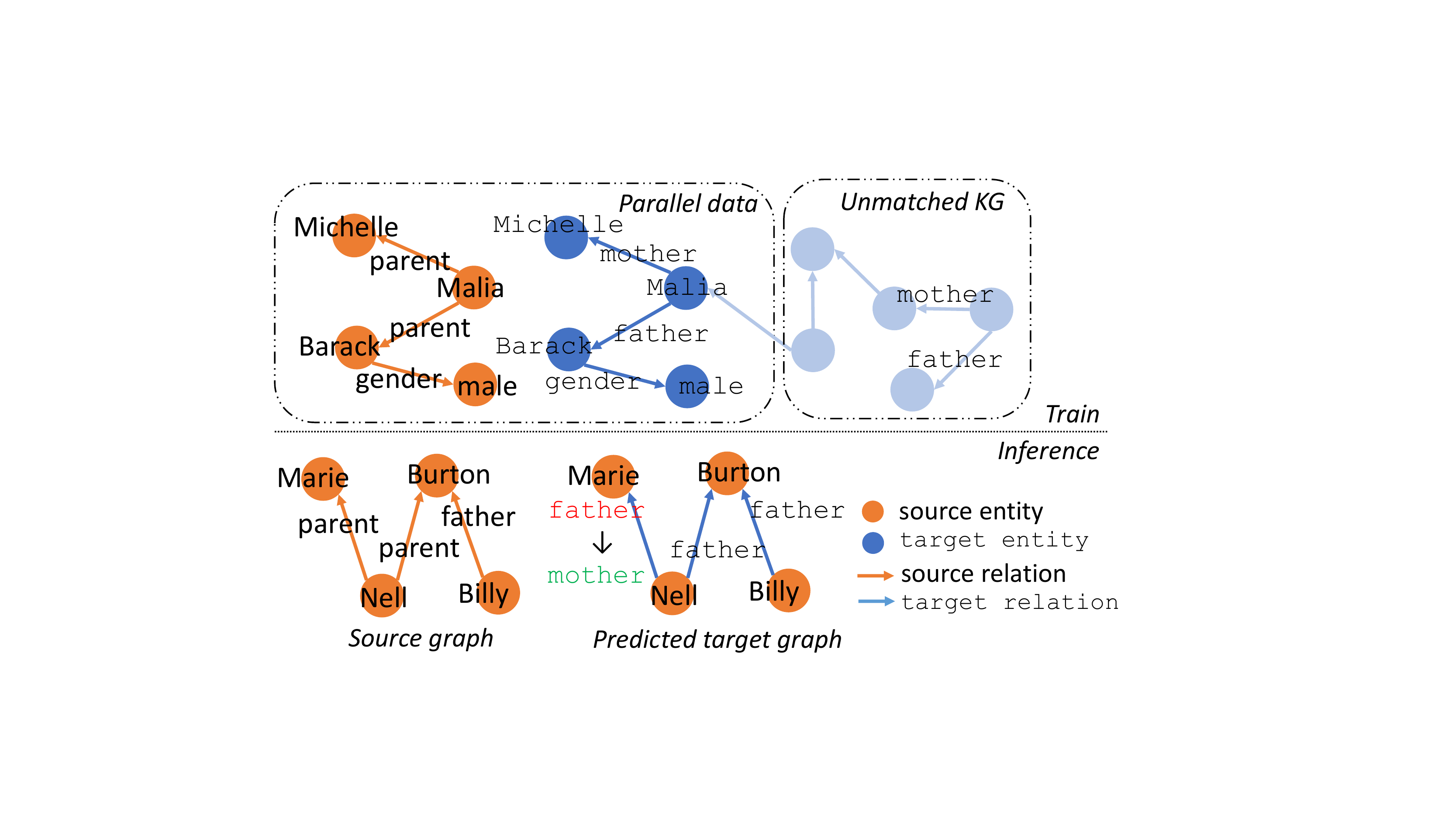}
\caption{A motivating example. Trained on \emph{parallel data}, a local model may suffer from sparse context for a new entity pair Nell-Marie at inference, wrongly disambiguating ``parent'' to \textcolor{red}{\texttt{father}} instead of \textcolor{green}{\texttt{mother}}.}
\label{fig:overall}
\end{figure}

\paratitle{Local Approaches.} Relation integration has been studied by the natural language processing (NLP) community.
With exact matching in literal form between entity names in the source graph and target KG, previous methods obtain \emph{parallel data}, \ie common entity pairs, between the two graphs as in \autoref{fig:overall}.
Features of the entity pairs (\eg Malia-Barack) in the source graph and their relations in the target KG (\eg \texttt{father}) are used to train models to predict target relations for future extractions.
A common challenge is the \emph{ambiguity} of source relations, \eg ``parent'' may correspond to \texttt{father} or \texttt{mother} in different contexts.
Previous methods exploited \emph{contextual} features including embeddings of seen entities (\eg ``Malia''; \citealt{riedel-2013-unischema}), middle relations between (\eg ``parent'';~\citealt{riedel-2013-unischema,toutanova-2015-joint,verga-2017-rowless,verga-2016-columnless,weston-2013-kbreemb}), and neighbor relations around the entity pair (\eg ``gender'';~\citealt{,zhang-2019-openki}).

Assuming rich contexts to address the ambiguity challenge, previous methods may fall short under the evolving and incomplete nature of the source graph.
For example, in the lower part of \autoref{fig:overall}, emerging entities may come from new extractions with sparse contextual information.
For the pair Nell-Marie, a conventional model learned on the parallel data may have neither seen entities nor neighborhood information (\eg ``gender'') to depend on, thus failing to disambiguate ``parent'' and wrongly predicting \textcolor{red}{\texttt{father}}.
Due to the \emph{local} nature of previous approaches, \ie predictions for different entity pairs are made independently of each other, the model is unaware that ``Nell'' has two fathers in the final predictions.
Such predictions are incoherent in common sense that a person is more likely to have one father and one mother, which is indicated by the graph structure around \texttt{Malia} in the target KG part of the parallel data.

\subsection{Our Collective Approach}\label{ssec:collective}
To alleviate the incoherent prediction issue of local approaches, we propose \textbf{Co}llective \textbf{R}elation \textbf{I}ntegration (CoRI) that exploits the dependency of predictions between adjacent entity pairs to enforce global coherence.

Specifically, we follow two stages, \ie \emph{candidate generation} and \emph{collective inference}. 
In candidate generation, we simply use a local model to make independent predictions as candidates, \eg \texttt{father} for all the three pairs in the lower part of \autoref{fig:overall}.
In collective inference, we employ a \emph{collective} model that is aware of the common substructures of the target graph, \eg \texttt{Malia}.
The collective model makes predictions by not only taking as input all contextual features to the local model but also the candidate predictions of the current and all neighbor pairs.
For the pair Nell-Marie, the collective model will have access to the candidate prediction \texttt{father} of Nell-Burton, which helps flip its final prediction to the correct \textcolor{green}{\texttt{mother}}.
\autoref{tab:overview} summarizes CoRI and representative previous work from four aspects.
To the best of our knowledge, CoRI is the first to collectively perform relation integration rather than locally.

\begin{table}[t]
\setlength\tabcolsep{1.2pt}
\small
\centering
\begin{tabular}{lcccc}
\toprule
\multirow{2}{*}{\textbf{Methods}} &Middle &No entity & Neighbor& \textbf{Collective} \\
 & relation &param. & relation & \textbf{inference} \\
\midrule
\cite{riedel-2013-unischema} &\checkmark &  & & \\
\cite{verga-2017-rowless} & \checkmark & \checkmark & & \\
\cite{zhang-2019-openki} & \checkmark & \checkmark & \checkmark & \\
CoRI (ours)& \checkmark & \checkmark & \checkmark & \checkmark \\
\bottomrule
\end{tabular}
\caption{Comparisons between CoRI and baselines.}
\label{tab:overview}
\end{table}

Being responsible to make globally consistent predictions, the collective model needs to be trained to encode common structures of the target KG, \eg \texttt{Malia} having only one father/mother in the parallel data of \autoref{fig:overall}.
To this end, we train the collective model in a \emph{stacked} manner~\cite{wolpert-stacked-1992}.
We first train the first-stage local model on the parallel data, then train the second-stage collective model by conditioning on the candidate predictions of neighbor entity pairs from the first stage (\eg \texttt{father} for Malia-Barrack) to make globally consistent predictions (\eg \texttt{mother} for Malia-Michelle).

\paratitle{Parallel Data Augmentation.} 
The parallel data may be bounded by the low recall of exact name matching or the limited extractions generated by open IE systems.
We observe that, even without counterpart extractions, the \emph{\unused} part of the target graph (as in \autoref{fig:overall}) may also have rich common structures to guide the training of the collective model.
To this end, we propose \emph{augmenting} the parallel data by sampling subgraphs from the \unused KG and creating pseudo parallel data by synthesizing their extractions, so the collective model can benefit from additional training data characterizing the desired global coherence.

To summarize, our contributions are three-fold: \textbf{(1)} We propose CoRI, a two-stage framework that improves state-of-the-art methods by making collective predictions with global coherence.
\textbf{(2)} We propose using the \unused target KG to augment the training data.
\textbf{(3)} Experimental results on two datasets demonstrate the superiority of our approaches, improving AUC from .677 to .748 and from .716 to .780, respectively.

\section{Preliminaries}
In this section, we first formulate the task of relation integration, then describe local methods by exemplifying with the state-of-the-art approach OpenKI~\cite{zhang-2019-openki}.

\subsection{Relation Integration}
We treat subject-relation-object extractions from open IE systems as a \emph{source graph} $\mathcal{K}(\mathcal{E}, \mathcal{R})=\{(s,r,o)\mid s,o\in\mathcal{E},r\in\mathcal{R}\}$, where $\mathcal{E}$ denotes extracted textual entities, \eg ``Barack Obama'', and $\mathcal{R}$ denotes extracted source relations, \eg ``parent''.
We denote by $(s, o)$ a source entity pair.
For $(s, o)$, $\mathcal{K}_{s,o}=\{r \mid(s,r,o) \in \mathcal{K}\}$ denotes all source relations between them.
Similarly, $\mathcal{K}_r=\{s,o\mid(s,r,o) \in \mathcal{K}\}$ denotes all entity pairs with relation $r$ in between.
We use the union $\mathcal{K}_{\mathcal{R}}=\bigcup_{r\in \mathcal{R}}{\mathcal{K}_r}$ to refer to all extracted entity pairs.

\begin{definition}[\textbf{Relation Integration}]
Given a source graph $\mathcal{K}$ and a \emph{target KG} $\mathcal{K}'(\mathcal{E}', \mathcal{R}')$ with target entities $\mathcal{E}'$ and target relations $\mathcal{R}'$, the task of relation integration is to predict all applicable target relations for each extracted entity pair in $\mathcal{K}_{\mathcal{R}}$:
\begin{equation*}
\Gamma\subseteq \mathcal{K}_{\mathcal{R}}\times \mathcal{R}',
\end{equation*}
where $(s,r',o)\in \Gamma$ is an \emph{integrated extraction} indicating that a target relation $r'$ holds for $(s,o)$.
\end{definition}

To train relation integration models, all methods employ \emph{parallel data} formalized as follow:
\begin{definition}[\textbf{Parallel Data}]\label{def:parallelData}
Parallel data are common entity pairs shared between $\mathcal{K}_{\mathcal{R}}$ and $\mathcal{K}'_{\mathcal{R}'}$ and their ground truth target relations in $\mathcal{K}'$: $\mathcal{T}=\{\langle (s, o),\mathcal{K}'_{s,o}\rangle \mid (s, o) \in \mathcal{K}_{\mathcal{R}}\cap\mathcal{K}'_{\mathcal{R}'}\}$. 
For example, $\langle (\text{Malia}, \text{Barack}), \{\texttt{father}\}\rangle$ is an instance of parallel data in \autoref{fig:overall}.
\end{definition}
To obtain parallel data, a widely used approach is to find entities shared by $\mathcal{E}$ and $\mathcal{E}'$ by exact name matching, then generate common entity pairs and their ground truth.

\subsection{Local Approaches}\label{sec:ind}
Previous local methods score potential integrated extractions by assuming their independence:
\begin{equation}
P(\Gamma \mid \mathcal{K})= \prod_{(s,r',o) \in \mathcal{K}_{\mathcal{R}}\times \mathcal{R}'} P_\theta(s,r',o\mid \mathcal{K}),
\end{equation}
where $\theta$ is the parameters of the local model.
One representative local model achieving state-of-the-art performance is OpenKI~\cite{zhang-2019-openki}.
It encodes the neighborhood of $(s,o)$ in $\mathcal{K}$ by grouping and averaging embeddings of source relations in three parts.
Let $\mathcal{K}_{s,\cdot}$ be the set of source relations between $s$ and neighbor entities other than $o$, and similarly for $\mathcal{K}_{\cdot,o}$.
OpenKI represents $(s,o)$ by concatenating the three averaged embeddings into a \emph{local representation} $\mathbf{t}_l$:
\begin{equation}
\mathbf{t}_l=[A(\mathcal{K}_{s,o});A(\mathcal{K}_{s,\cdot});A(\mathcal{K}_{\cdot,o})],
\end{equation}
where $l$ stands for local, and $A(.)$ takes a set of relations and outputs the average of their embeddings.
Then each integrated extraction is scored by:
\begin{equation}\label{eq:r_embedding}
P_\theta(s,r',o\mid \mathcal{K}) = \sigma(\mathbf{MLP}_l(\mathbf{t}_l))_{r'},
\end{equation}
where $\mathbf{MLP}_l$ is a multi-layer perceptron and $\sigma$ the sigmoid function.
Given a parallel data $\mathcal{T}=\{\langle (s,o),\mathcal{K}'_{s,o}\rangle\}$, the loss function per training example trades between maximizing the probabilities of positive target relations and minimizing those of negative target relations:
\begin{align}\label{eq:loss}
&L\big((s,o), \mathcal{K}'_{s,o}\big) = -\frac{1}{|\mathcal{K}'_{s,o}|}\sum_{r' \in \mathcal{K}'_{s,o}}{\log P_\theta(s,r',o\mid \mathcal{K})} \nonumber\\
&+ \frac{\gamma }{|\mathcal{R}' \setminus\mathcal{K}'_{s,o}|}\sum_{r' \in \mathcal{R}' \setminus\mathcal{K}'_{s,o}}{\log P_\theta(s,r',o\mid \mathcal{K})},
\end{align}
where $\gamma$ is a hyperparameter to account for the imbanlance between positive and negative relations, because the latter often outnumber the former.
The final loss is the sum of all examples.

\section{Collective Relation Integration}

As discussed in \autoref{sec:intro}, the drawback of local methods is that predictions of different entity pairs are independently made.
Neglecting their dependency may lead to predictions inconsistent with each other.

To address the issue, we propose a collective approach CoRI, which achieves collective relation integration via two stages: candidate generation and collective inference.
In this section, we demonstrate the input and output of the two stages, as well as our current implementations.

\begin{figure}
\centering
\includegraphics[width=0.95\columnwidth, clip, keepaspectratio]{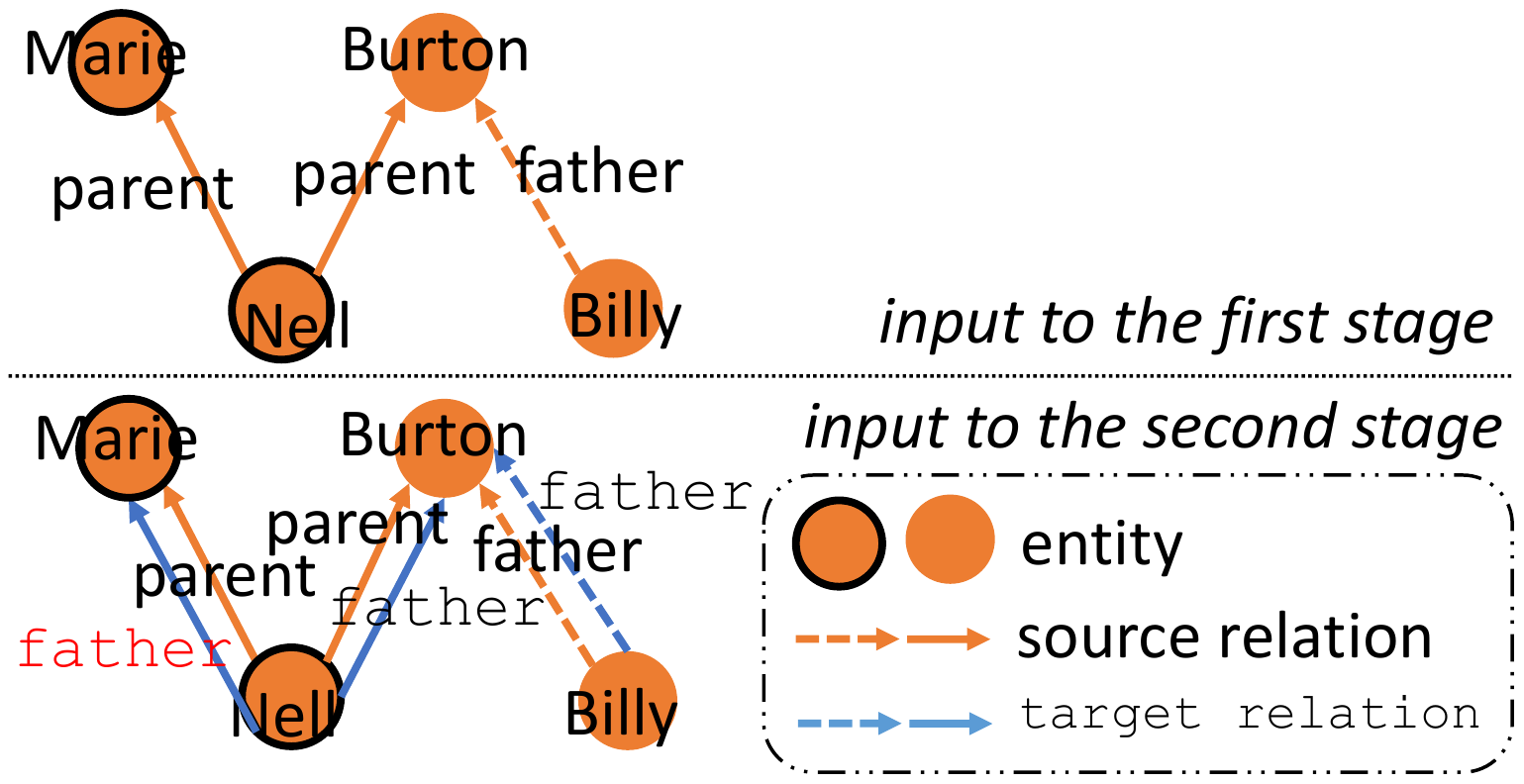}
\caption{Input of both stages on the Nell-Marie case. Solid edges are features for Nell-Marie. \textcolor{blue}{Additional edges} in the lower part are predicted candidate target relations $\Gamma^l$.}
\label{fig:twostage}
\end{figure}

\subsection{Candidate Generation}

As mentioned in \autoref{ssec:collective}, candidate generation's responsibility is to provide candidate predictions to the collective inference stage.
Formally, candidate predictions $\Gamma^l$ ($l$ means local) are generated by executing a local model on the source graph $\mathcal{K}$:
\begin{equation}
\Gamma^l=\argmax_{\Gamma} P_\theta(\Gamma \mid \mathcal{K}).
\end{equation}
The candidate predictions in $\Gamma^l$ may be partially wrong, but the other correct ones can help adjust wrong predictions of their adjacent entity pairs in the collective inference stage, under the guidance of the collective model.

For example, in the upper part of \autoref{fig:twostage}, we have a source graph $\mathcal{K}$ with three entity pairs.
The input to candidate generation is the entire $\mathcal{K}$.
After applying the local model (OpenKI in our case), we have three additional edges as the output $\Gamma^l$ in the lower part of \autoref{fig:twostage}.
Note that the candidate prediction \texttt{\textcolor{red}{father}} for Nell-Marie (denoted by black outline) is incorrect due to insufficient information in its neighborhood in $\mathcal{K}$, \ie both the relations in between of and around the entity pair (denoted by solid edges) are ambiguous ``parent''s.

Fortunately, the entity pair Nell-Burton is relatively easy for the local model to predict as \texttt{father} because it can leverage the neighbor relation ``father'' between Billy-Burton.
Such correct candidate predictions are included in $\Gamma^l$, provided to the collective inference stage as additional signals for later correction of the wrong predictions such as \texttt{\textcolor{red}{father}} for Nell-Marie.

\subsection{Collective Inference}

Collective inference's responsibility is to encode the structures of the target graph and use such information to refine the candidate predictions $\Gamma^l$ by enforcing coherence among them.
To this end, a collective model $P_\beta$ (with parameters $\beta$) takes both the source graph $\mathcal{K}$ and the candidate predictions $\Gamma^l$ as input, and outputs the final predictions $\Gamma$:
\begin{equation}\label{eq:cmodel}
P(\Gamma \mid \mathcal{K})= P_\beta(\Gamma\mid \mathcal{K},\Gamma^l).
\end{equation}

In the Nell-Marie case of \autoref{fig:twostage}, when making the final prediction, its own candidate predictions and those of the neighbor entity pairs (solid edges in $\Gamma^l$ of the lower part in \autoref{fig:twostage}) are used to leverage the dependency among them.
We concatenate the embeddings of candidate predictions to the local representation $\mathbf{t}_l$ obtained in the first stage, and represent each entity pair as follow:
\begin{equation}\label{eq:concatCandidate}
\mathbf{t}_c=[\mathbf{t}_l;A(\Gamma^l_{s,o});A(\Gamma^l_{s,\cdot});A(\Gamma^l_{\cdot,o})],
\end{equation}
where $c$ means collective.
$\Gamma^l_{s,o}$ includes candidate target relations between $s$ and $o$, and similarly for $\Gamma^l_{s,\cdot}$ and $\Gamma^l_{\cdot,o}$.
Then we use another multi-layer perceptron $\mathbf{MLP}_c$ to convert $\mathbf{t}_c$ to probabilities
\begin{equation}\label{eq:second_stage}
P_\beta(s,r',o\mid \mathcal{K},\Gamma^l) =\sigma(\mathbf{MLP}_c(\mathbf{t}_c))_{r'},
\end{equation}
and minimize the loss function for $P_\beta$ similar to that of the local model $P_\theta$ in \autoref{eq:loss}.

\subsection{Training Collective Model} 

\begin{algorithm}[t]
\small
\SetAlgoLined
\KwResult{Collective model $\beta$.}
$\mathcal{T}^{(1)}, \dots, \mathcal{T}^{(T)}\gets$ Split training data $\mathcal{T}$ into $T$ folds\;
\For{fold $i=1 , \dots, T$}{
$\theta^{(i)} \gets$ train local model on data folds $1, \dots, i-1, i+1, \dots, T$\;
$\Gamma^l_i \gets$ local predictions on $\mathcal{T}^{(i)}$ using $\theta^{(i)}$\;
}
$\Gamma^l \gets \cup_i \Gamma^l_i$\;
$\beta \gets$ train collective model on $\mathcal{T}$ with input $\mathcal{K}$ and $\Gamma^l$\;
\caption{Training collective model.\label{alg:overall}}
\end{algorithm}

According to \autoref{eq:cmodel}, we need $\Gamma^l$ as features to train the collective model $P_\beta$. 
This is to ensure that $P_\beta$ captures the dependencies among target relations.
One may ask why we do not directly use ground truth $\mathcal{K}'$ instead of predictions $\Gamma^l$.
At test time, we can only use target relations predicted by $P_\theta$ as input to $P_\beta$ because the ground truth target relations of neighbor entity pairs might not be available.
If we train $P_\beta$ using the ground truth, there will be a discrepancy between training and testing, potentially hurting the performance.

Specifically, we split the training set $\mathcal{T}$ into $T$ folds.
We generate $\Gamma^l$ by rotating and unioning a temporary local model's predictions on a held-out fold, where the temporary model is trained on the other folds.
Then we train $P_\beta$ on the parallel data $\mathcal{T}$ with $\Gamma^l$.
In this manner, we can use the full dataset to optimize the collective model while avoiding generating candidates on the training data of the local model, which leads to overfitting.
The detailed training procedure is given in \autoref{alg:overall}. 

\section{Data Augmentation w/ Unmatched KG}\label{sec:untouch}

As in \autoref{def:parallelData}, the volume of parallel data is limited by the number of shared entity pairs $\mathcal{K}_{\mathcal{R}}\cap\mathcal{K}'_{\mathcal{R}'}$ of the two graphs.
In \autoref{fig:overall}, the \emph{\unused} part of the target KG, containing entity pairs without extraction counterparts (\ie $\mathcal{K}'_{\mathcal{R}'}\setminus \mathcal{K}_{\mathcal{R}}$) and their target relations, can also indicate common substructures of the target KG, and guide the training of the collective model.
To this end, we propose leveraging \unused KG to generate \emph{pseudo parallel data} to augment the limited training data.

\begin{figure}[t]
\centering
\includegraphics[width=0.8\columnwidth, clip, keepaspectratio]{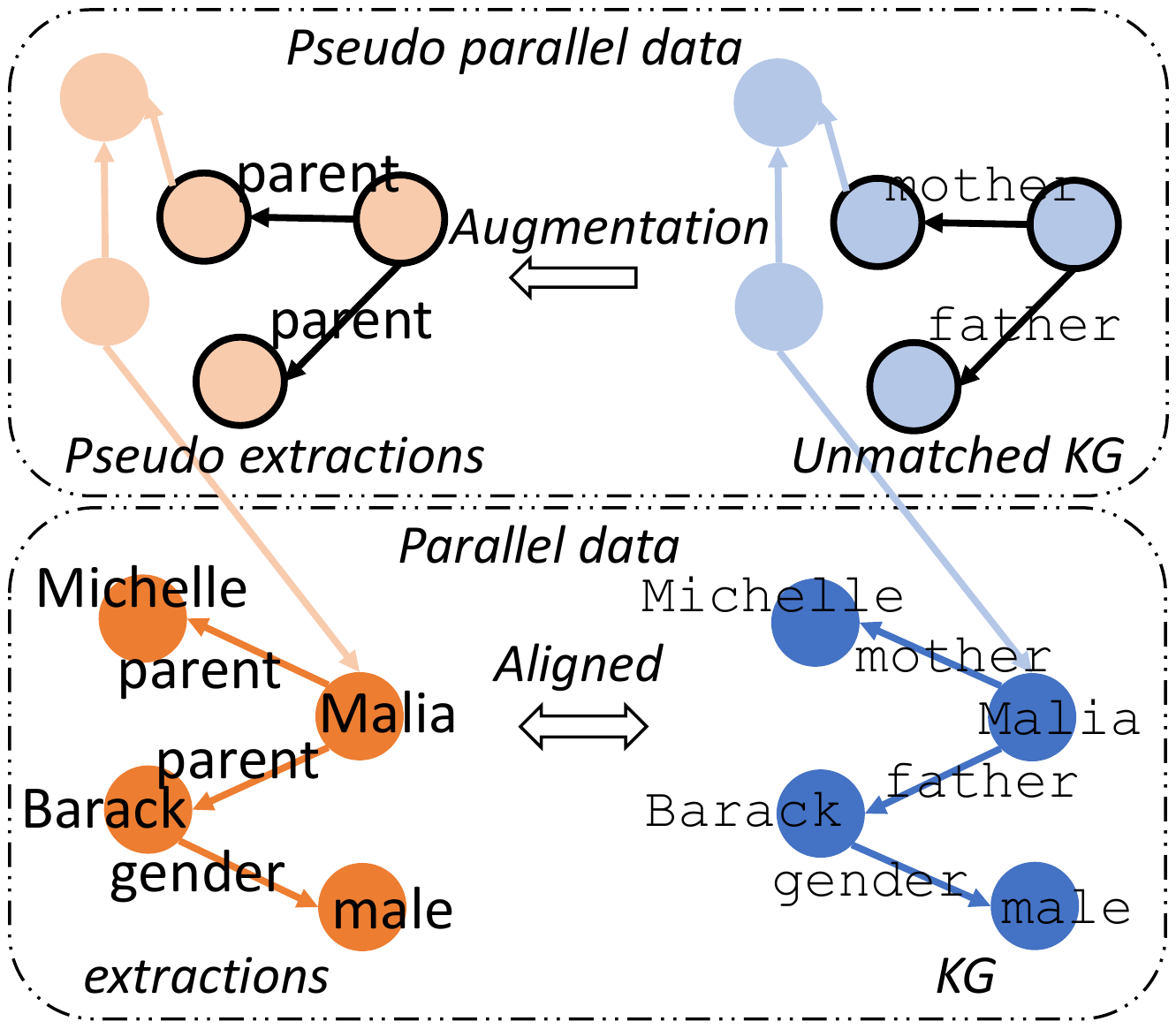}
\caption{Illustration of parallel data augmentation. We first generate pseudo extractions for the \unused KG, then select a subset of entity pairs that are similar to the parallel data (with black outline) to augment training.}
\label{fig:aug}
\end{figure}

\paratitle{Synthesizing Pseudo Extractions.}
To leverage the \unused KG, we need to synthesize \emph{pseudo extractions} for the target entities and relations to add to $\mathcal{K}$ as features.
Since we do not use entity-specific parameters, we only synthesize source relations like ``parent'', and keep the target entities unchanged, as illustrated in \autoref{fig:aug}.
Specifically, for each subject-relation-object tuple $(s', r', o')$ in the \unused KG, we keep $s'$ and $o'$ unchanged, and synthesize source relations $r$ by sampling from:
\begin{equation}\label{eq:translation}
P(r\mid r')=\frac{|\mathcal{K}_r\cap\mathcal{K}'_{r'}|}{|\mathcal{K}'_{r'}|},
\end{equation}
\ie the conditional probability of observing $r$ given $r'$ based on co-occurrences in the parallel data.
$|\mathcal{K}_r\cap\mathcal{K}'_{r'}|$ is the number of entity pairs with both $r$ and $r'$ in between, and $|\mathcal{K}'_{r'}|$ is the number of entity pairs with $r'$ in between.
In this way, we obtain a pseudo extraction $(s,r,o)$, as detailed in \autoref{alg:aug}

\paratitle{Pseudo Data Selection.}
We regard all pseudo extractions as a graph $\mathcal{K}^p$.
Similar to \autoref{def:parallelData}, we define \emph{pseudo parallel data} as below.
\begin{definition}[\textbf{Pseudo Parallel Data}]\label{def:pseudoData}
Pseudo parallel data $\mathcal{T}^p$ includes common entity pairs between pseudo extractions $\mathcal{K}^p$ and the target KG $\mathcal{K}'$, associated with their ground truth target relations, \ie $\mathcal{T}^p=\{\langle (s,o),\mathcal{K}'_{s,o}\rangle\} \mid (s,o)\in \mathcal{K}^p_{\mathcal{R}} \cap \mathcal{K}'_{\mathcal{R}'}\}$.
\end{definition}
To make use of pseudo parallel data $\mathcal{T}^p$, the most straightforward way is to use them together with parallel data $\mathcal{T}$ to train the collective model $P_\beta$.
However, not all substructures in the target graph $\mathcal{K}'$ are useful for $P_\beta$.
For example, when $\mathcal{K}'$ has other domains irrelevant to the source extraction graph, substructures in those domains may distract $P_\beta$ from concentrating on the domains of the source graph.
To mitigate this issue, we only use a subset of $\mathcal{T}^p$ similar to $\mathcal{T}$, as shown by the black-outlined parts in \autoref{fig:aug}.
Specifically, we represent each entity pair $(s,o)$ as a \emph{virtual document} with surrounding target relations $\mathcal{K}'_{s,o} \cup \mathcal{K}'_{s,\cdot} \cup \mathcal{K}'_{\cdot,o}$ as ``tokens''.
For each entity pair from the parallel data $\mathcal{T}$, we use BM25~\cite{rob-bm25} to retrieve its top $K$ most similar entity pairs from $\mathcal{T}^p$, and add them to the selected pseudo parallel data $\overline{\mathcal{T}^p}$ for training, as detailed in \autoref{alg:aug}.

\begin{algorithm}[t]
\small
\SetAlgoLined
\KwResult{Collective model $\beta$ with data augmentation.}
\textbf{(1) Synthesizing Pseudo Extractions $\mathcal{K}^p$} \\
$\mathcal{K}^{p} \gets \emptyset$; $\overline{\mathcal{T}^{p}} \gets \emptyset$;\\
\For{$(s',r',o')\in\mathcal{K}'$, where $(s',o') \in \mathcal{K}'_{\mathcal{R}'}\setminus \mathcal{K}_{\mathcal{R}}$}
{$s\gets s'$ and $o\gets o'$; \\
Sample $r\sim P(r|r')$;\\
$\mathcal{K}^p\gets \mathcal{K}^p\cup \{(s,r,o)\}$;}
\vspace{-2mm}
\noindent\rule{7cm}{0.5pt} \\
\textbf{(2) Pseudo Data Selection} \\
\For{entity pair $\in \mathcal{K}_{\mathcal{R}} \cap \mathcal{K}'_{\mathcal{R}'}$}{
$S \gets$ its top $K$ similar entity pairs in $\mathcal{K}^p_{\mathcal{R}} \cap \mathcal{K}'_{\mathcal{R}'}$;\\
$\overline{\mathcal{T}^{p}} \gets \overline{\mathcal{T}^{p}} \cup \{\langle (s,o),\mathcal{K}'_{s,o} \rangle\mid (s,o)\in S\}$;\\
}
$\beta \gets$ Train on $\mathcal{T}\cup \overline{\mathcal{T}^{p}}$ with \autoref{alg:overall};\\
\caption{Our augmentation approach.}\label{alg:aug}
\end{algorithm}

\section{Experimental Settings}

\subsection{Datasets and Evaluation}

We use the \textbf{ReVerb} dataset \cite{fader-2011-reverb} as the source graph, and \textbf{Freebase}\footnote{\url{https://developers.google.com/freebase}} and \textbf{Wikidata}\footnote{\url{https://www.wikidata.org}} as the target KGs, respectively.
We follow the same name matching approach in \citet{zhang-2019-openki} to obtain parallel data.
To simulate real scenarios where models are trained on limited labeled data but applied to a large testing set, we use 20\% of entity pairs in the parallel data for training and the other 80\% for testing, and there is \emph{no} overlap.
We also compare the performance under other ratios in \autoref{sec:datasize}.
Dataset statistics are listed in \autoref{tab:data}.

\begin{table}[h]
\small
\centering
\setlength\tabcolsep{8.5pt}
\begin{tabular}{lccc}
\toprule
\textbf{Datasets} & \textbf{\#Train} & \textbf{\#Test} & \textbf{$|\mathcal{R}|$} \\
\midrule
ReVerb + Freebase & 12,344 & 49,629 & 97,196 \\
ReVerb + Wikidata & 8,447 & 33,849 & 182,407 \\
\bottomrule
\end{tabular}
\caption{Dataset statistics. We follow \citet{zhang-2019-openki} to use the most frequent 250 target relations.}
\label{tab:data}
\end{table}

We evaluate by ranking all integrated extractions based on their probabilities, and report area under the curve (\textbf{AUC}).
Considering real scenarios where we want to integrate as many extractions as possible while keeping a high precision,
we also report \textbf{Recall} and \textbf{F$_1$} when precision is 0.8, 0.9, or 0.95.

\subsection{Compared Methods}
We compare the following methods in experiments.

\paratitle{Relation Translation} is a simple method that maps source relations to target relations with conditional probability $P(r' \mid r)$ similar to \autoref{eq:translation}.
For an entity pair $(s,o)$, the predicted target relations are $\{\arg\max_{r'} P(r'|r)\mid r\in \mathcal{K}_{s,o}\}$.

\begin{table*}[t]
\setlength\tabcolsep{6.3pt}
\small
\centering
\begin{tabular}{lcc@{\tinycol}cc@{\tinycol}cc@{\tinycol}c|cc@{\tinycol}cc@{\tinycol}cc@{\tinycol}c}
\toprule
\textbf{Datasets} & \multicolumn{7}{c|}{ReVerb + Freebase} & \multicolumn{7}{c}{ReVerb + Wikidata} \\
\midrule
\multirow{2}{*}{\textbf{Metrics}} & \multirow{2}{*}{AUC} & \multicolumn{2}{c}{Prec = 0.8} & \multicolumn{2}{c}{Prec = 0.9} &\multicolumn{2}{c|}{Prec = 0.95}& \multirow{2}{*}{AUC} & \multicolumn{2}{c}{Prec = 0.8} & \multicolumn{2}{c}{Prec = 0.9} &\multicolumn{2}{c}{Prec = 0.95}  \\
\cline{3-8}
\cline{10-15}
& & Rec&F$_1$ & Rec&F$_1$ & Rec&F$_1$ & & Rec&F$_1$ & Rec&F$_1$ & Rec&F$_1$  \\
\midrule
Translation & .571 & \underline{.590} & \underline{.679} & .100 &.180 & .067 &.125 & .604 & .595 &.683 & .088 &.160 & .042 &.080 \\
E-model & .205 & .014 &.027 & .010 &.020 & .005 &.010 & .214 & - & - & - & -& -& -\\
Rowless & .593 & .473 &.594 & .372 &.526 & .186 &.310 & .647 & .511 &.624 & .381 &.536 & .266 &.416 \\
OpenKI & \underline{.677} & .553 &.654 & \underline{.449} &\underline{.599} & \underline{.314} &\underline{.472} & \underline{.716} & \underline{.605} &\underline{.689} & \underline{.511} &\underline{.652} & \underline{.407} &\underline{.570} \\
\midrule
CoRI & .708 & .590 & .679 & .494 & .638 & .381 & .544 & .746 & .641 & .712 & .558 & .689 & .461 & .621 \\
\quad+ KGE & .711 & .597 & .684 & .514 & .654 & .418 & .581 & .763 & .662 & .725 & .596 & .717 & .520 & .672 \\
\quad+ DA (random) & .734 & .616 & .696 & .518 & .658 & .395 & .558 & .774 & .678 & .734 & .606 & .724 & .521 & .673 \\
\quad+ DA (retrieval) & \textbf{.748} & \textbf{.636} &\textbf{.708} & \textbf{.539} &\textbf{.674} & \textbf{.421} &\textbf{.583} & \textbf{.780} & \textbf{.685} &\textbf{.738} & \textbf{.613} &\textbf{.729} & \textbf{.529} &\textbf{.680} \\
\bottomrule
\end{tabular}
\caption{Main experimental results. The best results are \textbf{in bold}, and the best external baselines are \underline{underlined}.
CoRI outperforms the best baseline OpenKI by a large margin, and parallel data augmentation (DA) further improves its performance. ``-'' indicates that the precision was not achieved.}
\label{tab:overall}
\end{table*}

\paratitle{Universal Schema (E-model)}~\cite{riedel-2013-unischema} learns entity and relation embeddings through matrix factorization, which cannot generalize to unseen entities.
It is a local model that scores each integrated extraction independently.

\paratitle{Rowless Universal Schema}~\cite{verga-2017-rowless} is a local model which improves over the E-model by eliminating entity-specific parameters, thus generalizing to unseen entities.

\paratitle{OpenKI}~\cite{zhang-2019-openki} is a local model that addresses the ambiguity of source relations by using neighbor relations for more context.

\paratitle{CoRI} is our collective two-stage relation integration model trained with \autoref{alg:overall}. 

\paratitle{CoRI + DA} is our model where the training data is augmented by pseudo parallel data with \autoref{alg:aug}.
To verify the necessity of \textbf{retrieval}-based pseudo data selection, we also compare with a \textbf{random} DA baseline where we select $K$ random entity pairs.

\paratitle{CoRI + KGE} is another approach to exploit the \unused KG with KG embeddings (\textbf{KGE}) trained on the entire target KG in an unsupervised manner.
We initialize the embeddings of target relations averaged by $A(.)$ in \autoref{eq:concatCandidate} with TransE~\cite{bordes-2013-transe} embeddings trained on the target graph.

\subsection{Implementation Details}
We uniformly use 32-dimension embeddings for all relations, and AdamW~\cite{loshchilov-2019-adamw} optimizer with learning rate 0.01 and epsilon 10\textsuperscript{-8}.
The ratio $\gamma$ in \autoref{eq:loss} is set to 10.
We sample at most 30 neighbor source relations to handle entity pairs with too many neighbor relations.
We use $T=5$ folds in \autoref{alg:overall} to train our collective model.
We retrieve top $K=5$ entity pairs in pseudo data selection, adding about 20K and 12K entity pairs to the two datasets in \autoref{tab:data}, respectively.
We use BM25~\cite{rob-bm25} implementation in ElasticSearch\footnote{\url{https://www.elastic.co/}} in pseudo data selection.
We use the KGE released by OpenKE.\footnote{\url{https://github.com/thunlp/OpenKE}}
Our model is trained with 32 CPU cores and a single 2080Ti GPU, and it takes 1-2 hours to converge.

\section{Experimental Results}

We aim to answer the following questions:
\textbf{(1)} Is CoRI superior to local models?
\textbf{(2)} Is CoRI robust \wrt varying size of training and testing data?
\textbf{(3)} Is \unused KG useful for CoRI? Is our parallel data augmentation approach the best choice?

\subsection{Main Results}

\begin{figure}[tb]
\subfloat[ReVerb + Freebase]{
\includegraphics[width=0.53\columnwidth, clip, keepaspectratio]{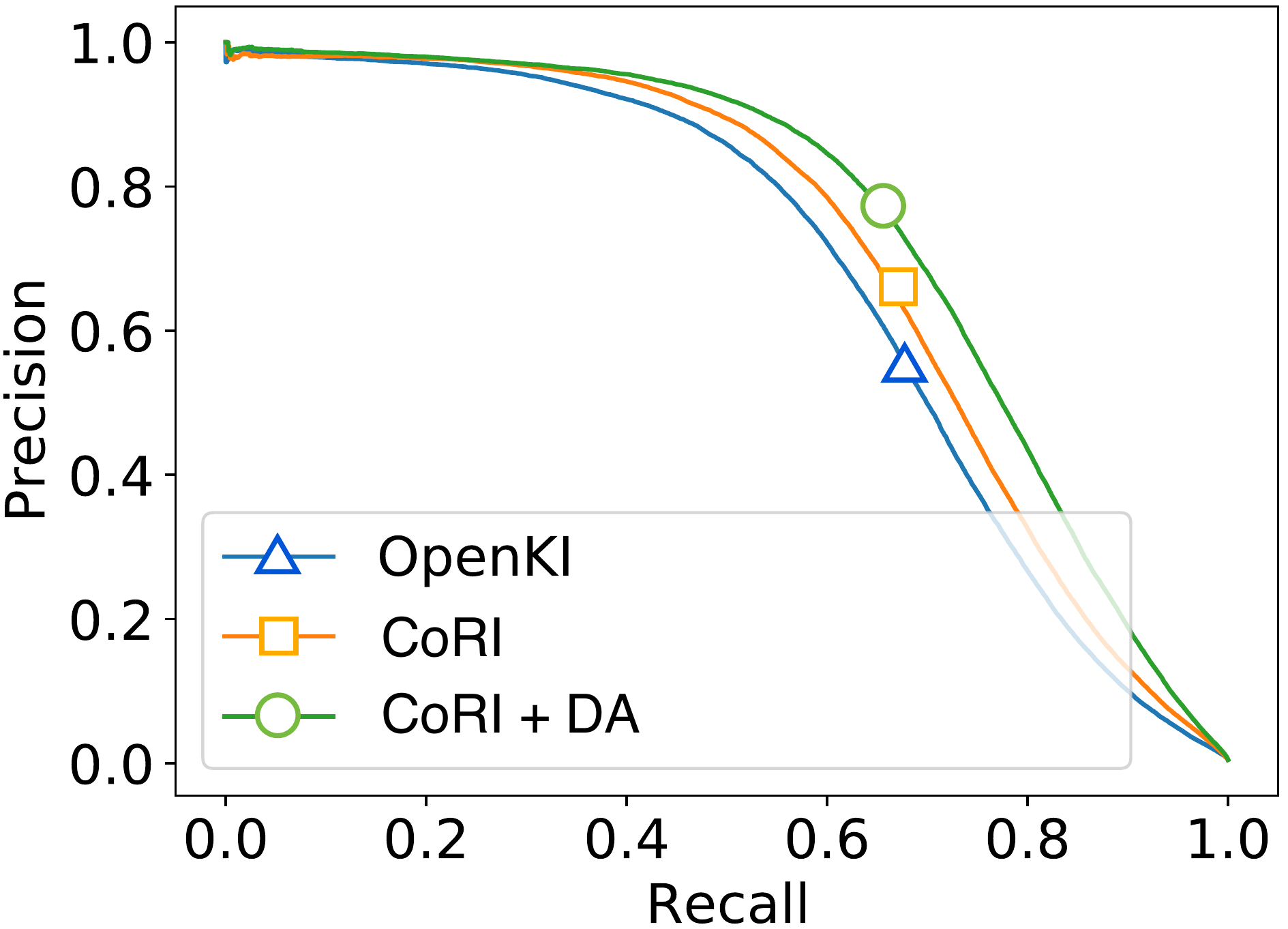}}
\subfloat[ReVerb + Wikidata]{
\includegraphics[width=0.462\columnwidth, clip, keepaspectratio]{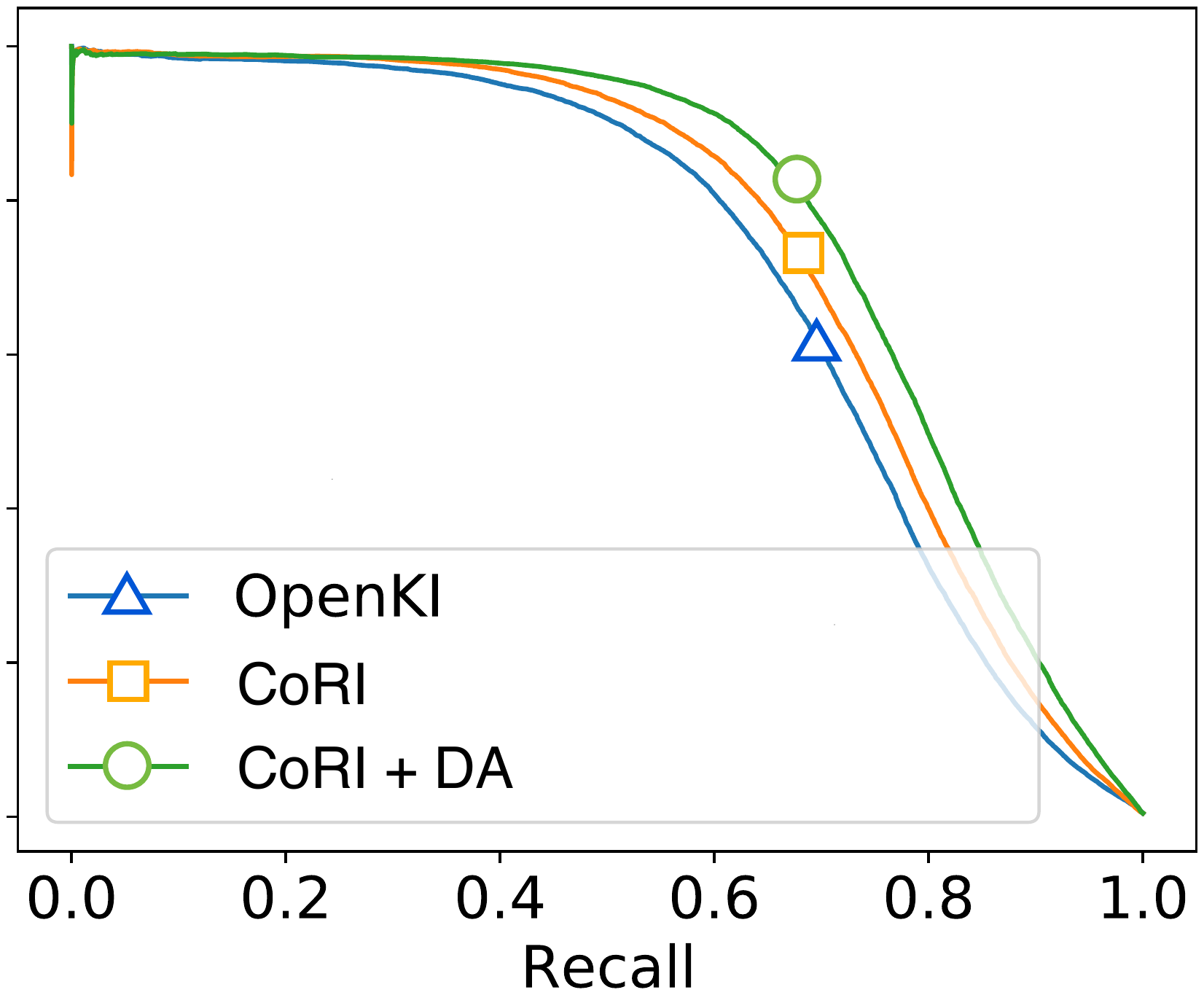}}
\caption{Precision-recall curves of best three methods.}
\label{fig:prcurve}
\end{figure} 
In \autoref{tab:overall}, we show results comparing all methods on both datasets.
Our observations are as follows.

\paratitle{Collective inference is beneficial.}
Among the baselines, OpenKI generally performs best because it leverages neighbor relations besides middle relations between entity pairs, without relying on entity parameters.
Even without data augmentation, CoRI outperforms OpenKI by a large margin, improving AUC from .677 to .708 and from .716 to .746 on the two datasets, respectively, which demonstrates the effectiveness of collective inference.

\paratitle{Data augmentation further improves the performance.}
By comparing CoRI with CoRI + DA (retrieval), we observe that data augmentation further improves AUC from .708 to .748 and from .746 to .780, respectively, which indicates that using \unused KG can effectively augment the training of the collective model.
We plot the precision-recall curves of the best three approaches in \autoref{fig:prcurve}.
It demonstrates the superiority of our methods across the whole spectrum.

\paratitle{Generalization on unseen entities is necessary.}
Among the baselines, the E-model uses entity-specific parameters, hindering it from generalizing to unseen entities and making it less competitive.

\subsection{Effectiveness of Pseudo Data Selection}

As shown in \autoref{tab:overall}, both KGE, random, and retrieval-based data augmentation approaches perform better than CoRI (without DA), indicating the effectiveness of using the \unused KG.
Our retrieval-based DA outperforms the random counterpart, which confirms the superiority of similarity-based data augmentation in choosing substructures that cover domains relevant to the original parallel data.
Our DA approach outperforms KGE, demonstrating the necessity of selectively using the unused KG to avoid discrepancies with the parallel data.

\begin{figure*}[t]
\hspace{-2mm}
\parbox{.29\linewidth}{
	\includegraphics[width=0.65\columnwidth, clip, keepaspectratio]{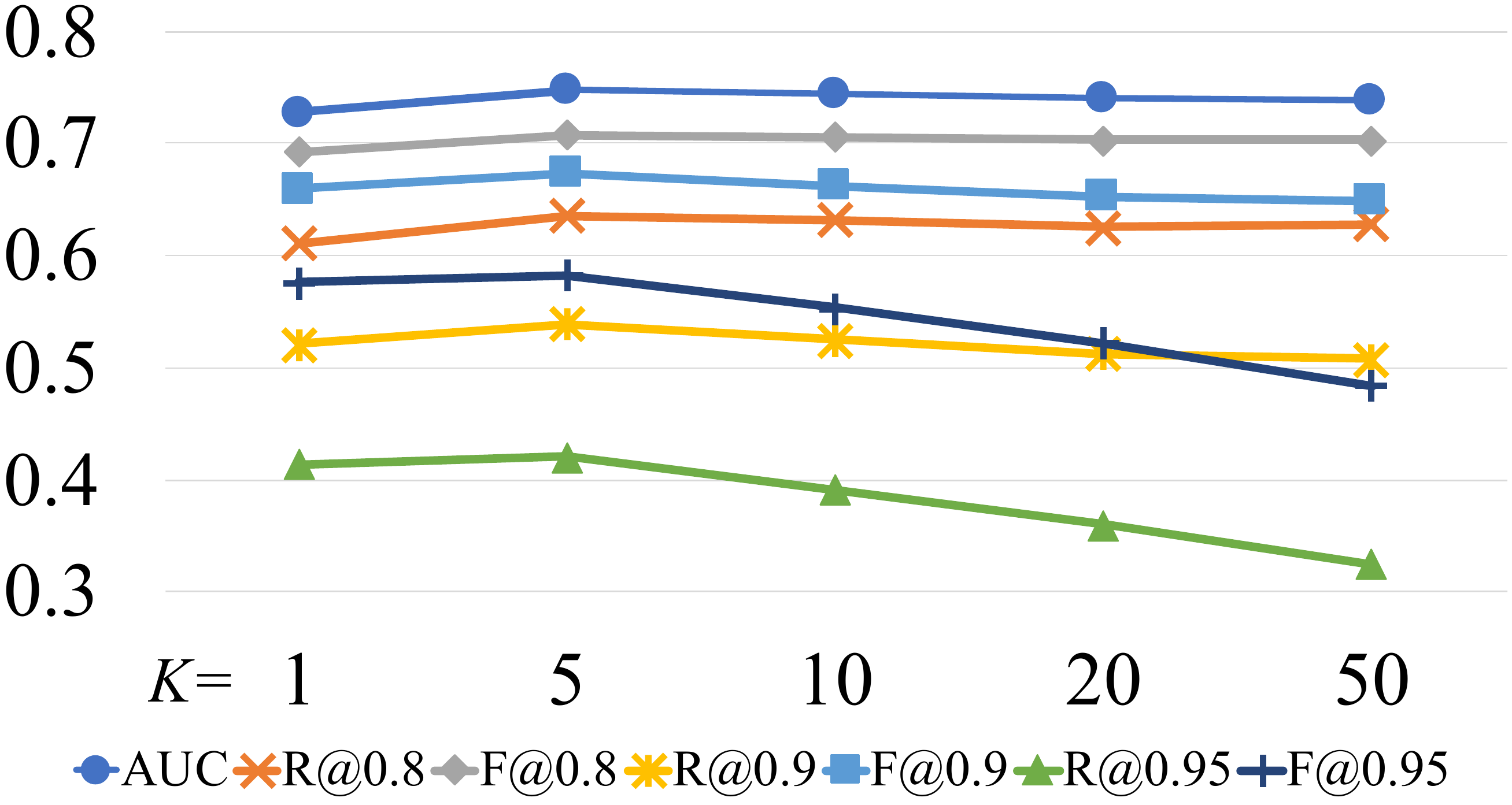}
	\vskip 1em
    \caption{Performance of data augmentation with different numbers of retrieved pairs $K$.}
    \label{fig:ret}
    }
    \hskip 0.7em
\parbox{.7\linewidth}{
    \subfloat[Varying size of training data.]{
     \includegraphics[width=0.72\columnwidth, clip, keepaspectratio]{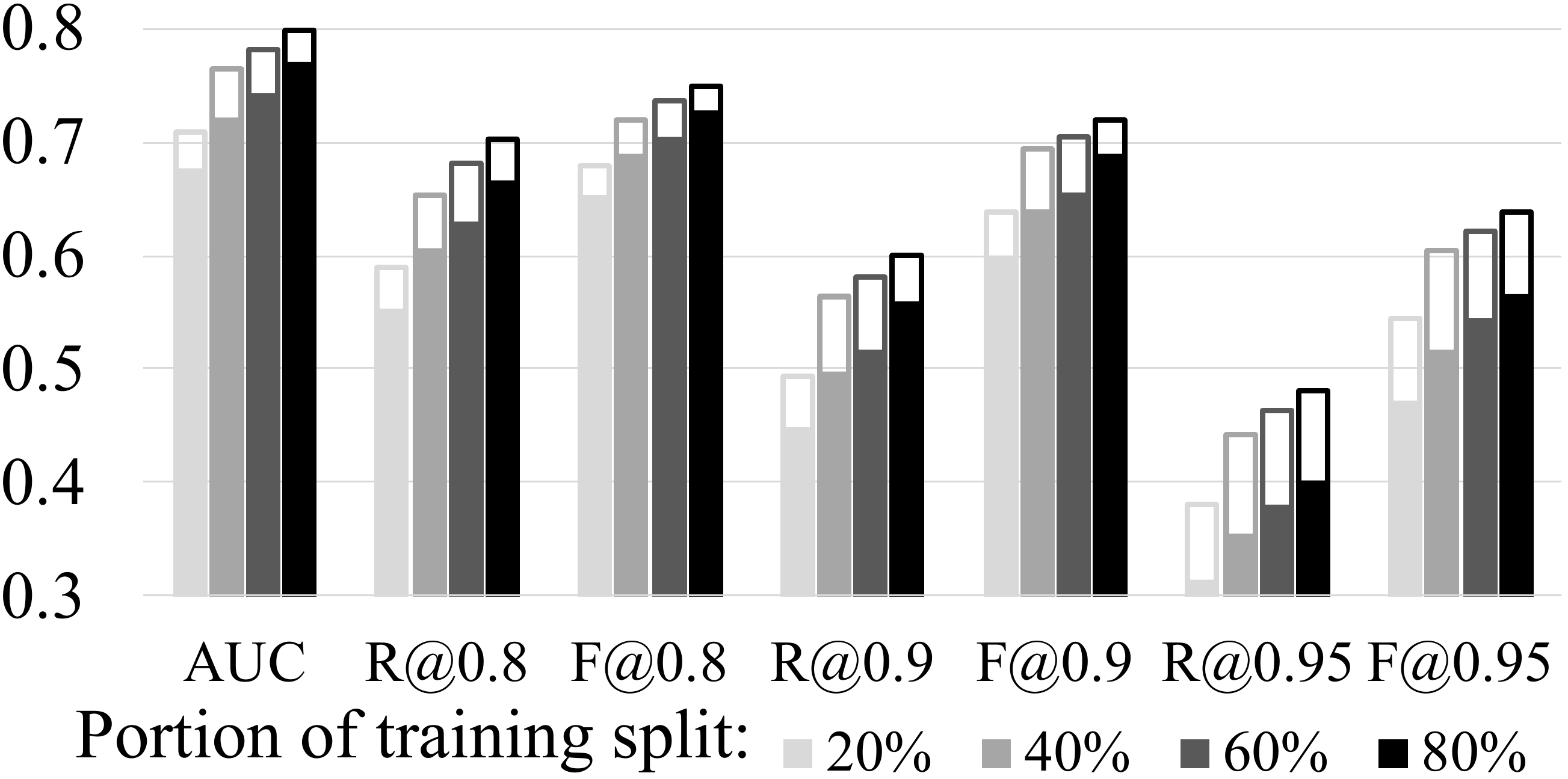}
        \label{fig:datasize}}
	\subfloat[Varying \% of accessible neighbor pairs]{
	\includegraphics[width=0.72\columnwidth, clip, keepaspectratio]{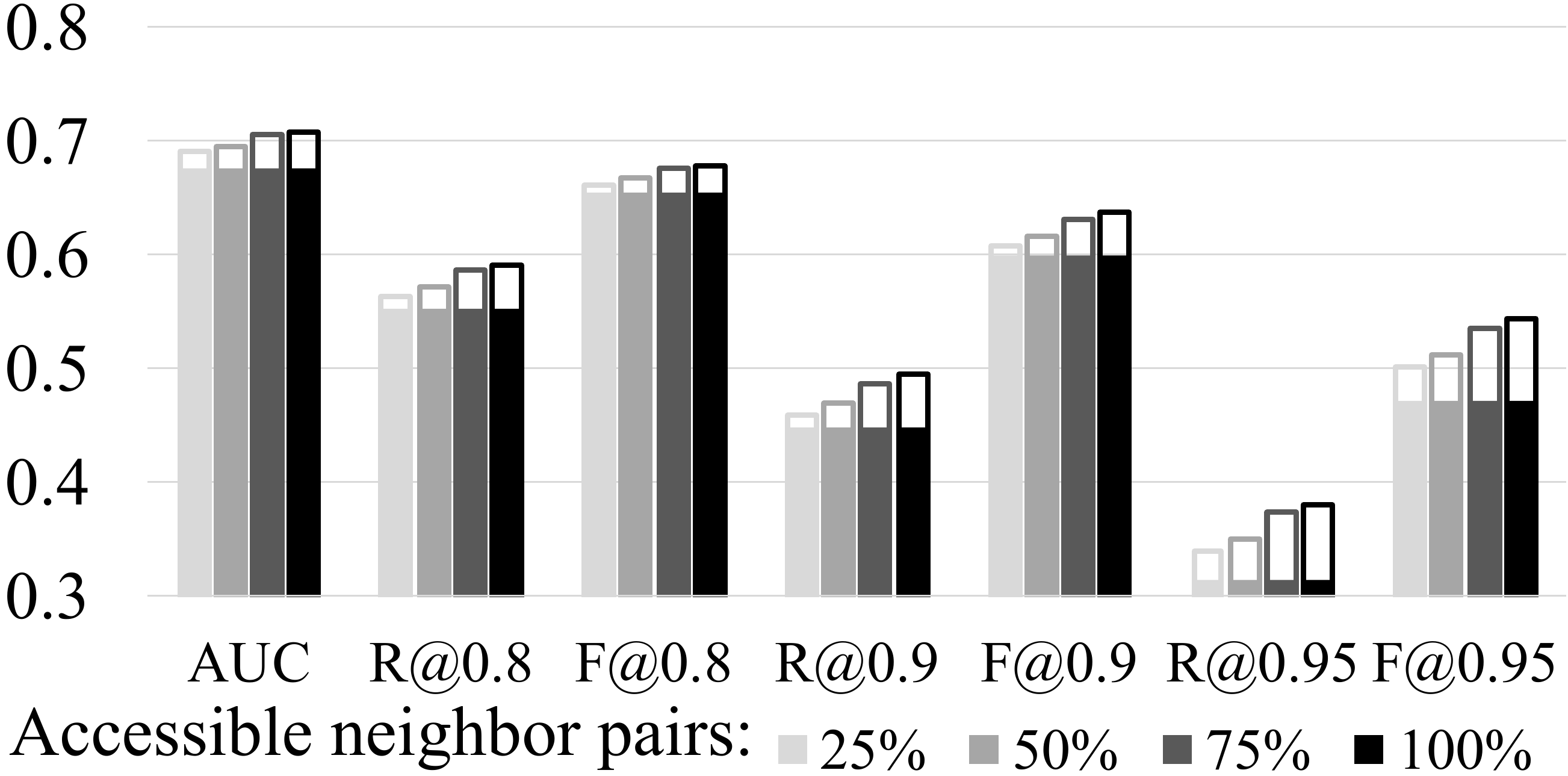}
    \label{fig:collective}
	}
	\caption{CoRI (bars without filling) vs. OpenKI (solid bars) on ReVerb + Freebase. CoRI consistently outperforms OpenKI by a large margin. Larger improvements are achieved when candidates of more neighbor entity pairs are accessed.}
    }
\end{figure*}

\paratitle{Different Numbers of Pseudo Data Entity Pairs.}\label{sec:ret}
In \autoref{fig:ret}, we compare the performance of DA \wrt different numbers of retrieved entity pairs $K$.
We observe that $K$=5 yields better performance than $K$=1.
However, further increasing $K$ hurts the performance, which is probably due to pseudo entity pairs with lower similarity to the parallel data causing a domain shift.
This validates the necessity of selectively using pseudo parallel data.

\subsection{Impacts of Data Size on CoRI}
Due to its collective nature, one may wonder about CoRI's performance \wrt other training and testing data sizes.
We analyze these factors in this section.
Our observations are similar on both datasets, so we only report the results on ReVerb + Freebase.

\paratitle{Varying Size of Training Data.}\label{sec:datasize}
In \autoref{fig:datasize}, we compare CoRI (without DA) with OpenKI by varying the portion of the parallel data for training from 20\% (used in our main results in \autoref{tab:overall}) to 80\%.
We observe that using more training data improves the performance, as shown by the increasing trends \wrt all metrics.
Our method outperforms OpenKI in all settings, demonstrating that our method is effective in both high- and low-resource settings.

\paratitle{Varying \% of Accessible Neighbor Entity Pairs.}
Our collective framework is special in its collective inference stage, where the collective model refines the candidate prediction of an entity pair by considering its neighbor entity pairs' candidates.
We hypothesize that the more neighbor entity pairs the collective model has access to, the better performance it should achieve.
For example, if we use a portion of 50\%, candidate predictions for only half of the neighbor entity pairs rather than the entire $\Gamma^l$ will be used in \autoref{eq:concatCandidate}.
We vary the portion from 25\% to 100\% (used in our main experiments in \autoref{tab:overall}).
As shown in \autoref{fig:collective}, even accessing 25\% can make CoRI outperform OpenKI.
As the percentage increases, CoRI continues to improve, while OpenKI remains the same because it is local, \ie not using candidate predictions.

\subsection{Case Study}
\begin{figure}[tb]
\centering
\includegraphics[width=1.0\columnwidth, clip, keepaspectratio]{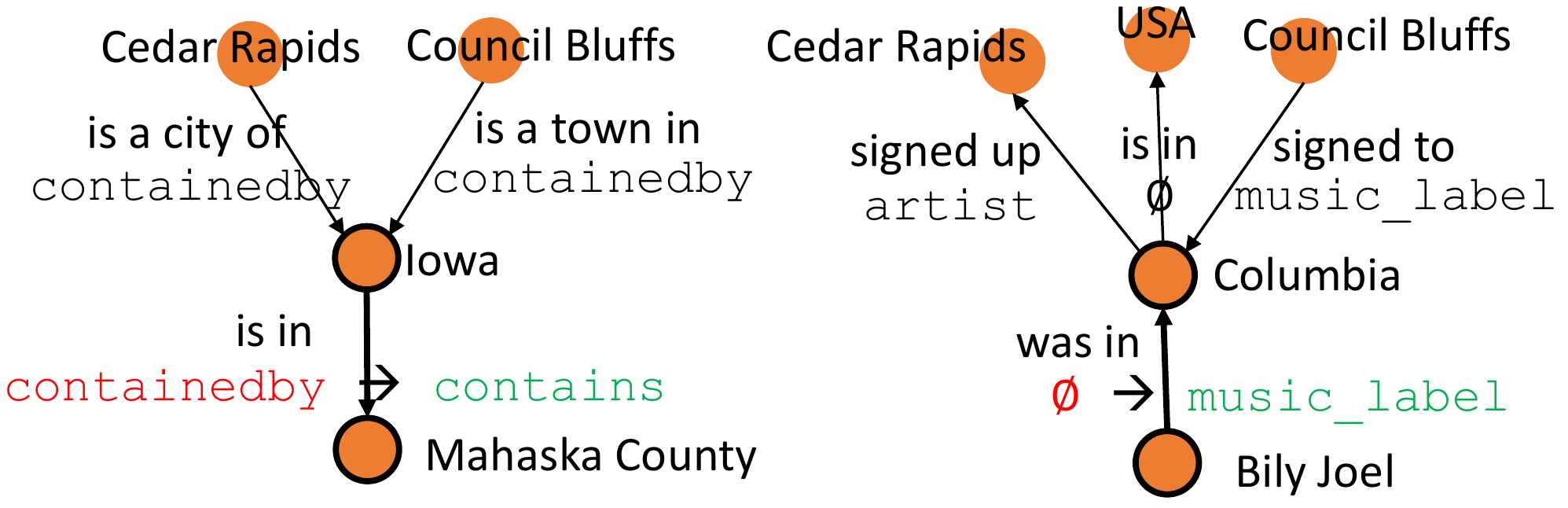}
\caption{Two cases from ReVerb + Freebase with predictions in this \texttt{font}. The \textcolor{red}{\texttt{wrong}} predictions of OpenKI is \textcolor{green}{\texttt{corrected}} by our collective model.}
\label{fig:case}
\end{figure}

In \autoref{fig:case}, we show two cases from ReVerb + Freebase where CoRI corrects the mistakes of OpenKI in the collective inference stage.
In the first case, the source relation ``is in'' between ``Iowa'' and ``Mahaska County'' is extracted but in the wrong direction.
OpenKI just straightforwardly predicts \texttt{containedby} based on the surface form, but fails to leverage the neighbor relations to infer that Iowa is a larger geographical area.
With the collective model, CoRI is able to use the other two candidate predictions of \texttt{containedby} to flip the wrong prediction to \texttt{contains}.

In the second case, a prediction is needed between ``Bily Joel'' and ``Columbia''.
Here the source relation ``was in'' and the object entity ``Columbia'' are both ambiguous, which can refer to geographical containment with a place or membership to a company.
OpenKI makes no prediction due to the ambiguity, while CoRI makes the right prediction \texttt{music\_label} by collectively working on the other entity pairs, where all predictions coherently indicate that ``Columbia'' is a music company.

\section{Related Work}

Relation integration has been studied by both the database (DB) and the NLP communities.
The DB community formulates it as schema matching that aligns the schemas of two tables, \eg matching columns of an \texttt{is\_in} table to those of another \texttt{subarea\_of} table~\cite{rahm-2001-schemamatching,cafarella-2008-webtables,kimming-2017-psl}.
Such table-level alignment is valid since all rows in an \texttt{is\_in} table should have the same semantics, \ie being geographical containment or not. 
However, in open IE, predictions should be made at the entity pair level because of the ambiguous nature of source relations.
Putting all extracted ``is in'' entity pairs into one table to conduct schema matching is problematic from the first step since the entity pairs may have different ground truths.

The NLP community, on the other hand, investigates the problem at the entity pair level. 
Besides manually designed rules \cite{soderland-2013-3hour}, most works leverage the link structure between entities and relations.
Universal schema~\cite{riedel-2013-unischema} learns embeddings of entities and middle relations between entity pairs through decomposing their co-occurrence matrix.
However, the entity embeddings make it not generalize to unseen entities.
Other methods~\cite{toutanova-2015-joint,verga-2016-columnless,verga-2017-rowless,gupta-2019-care} also exploit middle relations, but eliminate entity parameters.
\citet{zhang-2019-openki} moves one step further by explicitly considering neighbor relations, leveraging more context from the local link structure.
Some works~\cite{weston-2013-kbreemb,angeli-2015-stanfordoie} directly minimize the distance between embeddings of relations sharing the same entity pairs.
\citet{yu-2017-oie} further leverage compositional representations of entity names instead of using free parameters to deal with unseen entities at test time.

There are also works on Open IE canonicalization that cluster source relations.
Some use entity pairs as clustering signals~\cite{yates-2009-relsyn,nakashole-2012-patty,galarraga-coie-2014}, while others use lexical features or side information~\cite{min-2012-semanticre,vashishth-2018-cesi}.
However, the clusters are not finally aligned to relations in target KGs, different from our problem.

The two-stage collective inference framework has been explored in other problems like entity linking~\cite{cucerzan2007large,guo2013link,shen2012linden}, where candidate entities are generated for each mention independently, and collectively ranked based on their compatibility in the second stage. 
In machine translation, an effective approach to leverage monolingual corpus in the target language is to back-translate it to the source language to augment the limited parallel corpus~\cite{sennrich-2016-bt}.
The above works inspired us to use collective inference for relation integration and leverage the \unused KG for data augmentation.
Another approach to perform collective inference is to solve learning problem with constraints, such as integer linear programming \cite{roth-2014-lp}, posterior regularization \cite{ganchev-2010-pr}, and conditional random fields \cite{lafferty-2001-crf}.
Comparing to our approach, these methods usually involve heavy computation, or are hard to optimize.
Examining the performance of these methods is an interesting future direction.
Besides, we also adopted ideas of selecting samples from out-domain data similar to in-domain samples~\cite{xu-2020-extcode,du-2020-selfpretrain} to select our pseudo parallel data.

\section{Conclusion}
In this paper, we proposed CoRI, a collective inference approach to relation integration.
To the best of our knowledge, this is the first work exploring this idea.
We devised a two-stage framework, where the candidate generation stage employs existing local models to make candidate predictions, and the collective inference stage refines the candidate predictions by enforcing global coherence. 
Observing that the target KG is rich in substructures indicating the desired global coherence, we further proposed exploiting the \unused KG by selectively synthesizing pseudo parallel data to augment the training of our collective model.
Our solution significantly outperforms all baselines on two datasets, indicating the effectiveness of our approaches.

\section*{Acknowledgments}
We would like to thank Prashant Shiralkar, Hao Wei, Colin Lockard, Binxuan Huang, and all the reviewers for their insightful comments and suggestions.

\bibliographystyle{acl_natbib}
\bibliography{acl2021}

\end{document}